\documentclass[twoside,11pt]{article}

\usepackage[nohyperref]{jmlr2e}
\usepackage{amsmath,amssymb,amsfonts,bm} 	% Math stuff
\usepackage{verbatim}
\usepackage{framed}

\newcommand{\qu}[1]{``#1''}
\newcommand{\Set}[1]{\mathcal{#1}}
\def\N{\mathrm{I\kern-0.4ex N}}
\def\R{\mathrm{I\kern-0.4ex R}}
\def\E{\mathrm{I\kern-0.4ex E}}
\newcommand{\mat}[1]{\mathbf{#1}}
 
\newcommand{\op}[1]{{\rm \hat #1}}
\newcommand{\policyop}[1]{\op \Gamma{\kern-0.5ex_{#1}}}

\DeclareMathOperator*{\argmax}{ \text{\rm arg}\,\text{\rm max} }

\newcommand{\smallsum}[2]{ {\textstyle% 
	\sum\limits_{\scriptscriptstyle #1}^{\scriptscriptstyle #2}}% 
}

\newcommand{\smallfrac}[2]{ {\textstyle \frac{#1}{#2}} }

\ShortHeadings{Non-Deterministic Policy Improvement
	Stabilizes Approximated RL}{B\"ohmer, Guo and Obermayer}
\firstpageno{1}

\begin{document}
\thispagestyle{plain}

\begin{framed}
	\noindent
	{This paper has been presented at the 
	{\em 13th European Workshop on Reinforcement Learning} 
	({\bf EWRL 2016}) on the 3rd and 4th of December 2016
	in Barcelona, Spain.}
\end{framed}

\title{Non-Deterministic Policy Improvement \\
Stabilizes Approximated Reinforcement Learning}

\author{Wendelin B\"ohmer$^*$ \and Rong Guo \and Klaus Obermayer \\
		\addr{Neural Information Processing Group,
			Technische Universit\"at Berlin,}\\ 
		\addr{Marchstra\ss e 23, D-10587 Berlin, Germany.} \\
		{$^*$ corresponding author (email: {\sc wendelin@ni.tu-berlin.de})} }

\editor{Gergely Neu, Vince\c c G\'omez and Csaba Szepesv\'ari}

\maketitle

\begin{abstract}%   <- trailing '%' for backward compatibility of .sty file
	This paper investigates a type of instability 
	that is linked to the greedy policy improvement
	in approximated reinforcement learning. 
	We show empirically that non-deterministic policy improvement 
	can stabilize methods like LSPI by controlling 
	the improvements' stochasticity. 
	Additionally we show that a suitable representation 
	of the value function also stabilizes the solution to some degree.
	The presented approach is simple and should also 
	be easily transferable to more sophisticated algorithms
	like deep reinforcement learning.
\end{abstract}

%\vspace{4mm}

% ==============================================================================
\section{Introduction}

This paper investigates a type of instability
that is linked to the greedy policy improvement 
in approximated reinforcement learning.
We show empirically that non-deterministic policy improvement 
can be used to achieve stability for large discount factors.
The presented approach is simple and should also be 
easily transferable to more sophisticated algorithms.

Recently {\em deep reinforcement learning} (deep RL) has been very successful 
in solving complex tasks in large, often continuous state spaces
\citep[e.g.~playing Atari games and Go,][]{Mnih15,Silver16}.
These approaches use gradient based Q-learning \citep{Watkins92}
or policy gradient methods \citep{Williams92}.
Gradients in neural networks must be based 
on i.i.d.~distributed samples, though \citep[see][]{Riedmiller05}.
Deep RL uses therefore mini-batches that are sampled i.i.d.~from 
a fixed set of experiences, which has been collected before training
\citep[called {\em experience replay},][]{Mnih13}.

\begin{figure}[t!]
	\includegraphics[width=0.485\textwidth]%
		{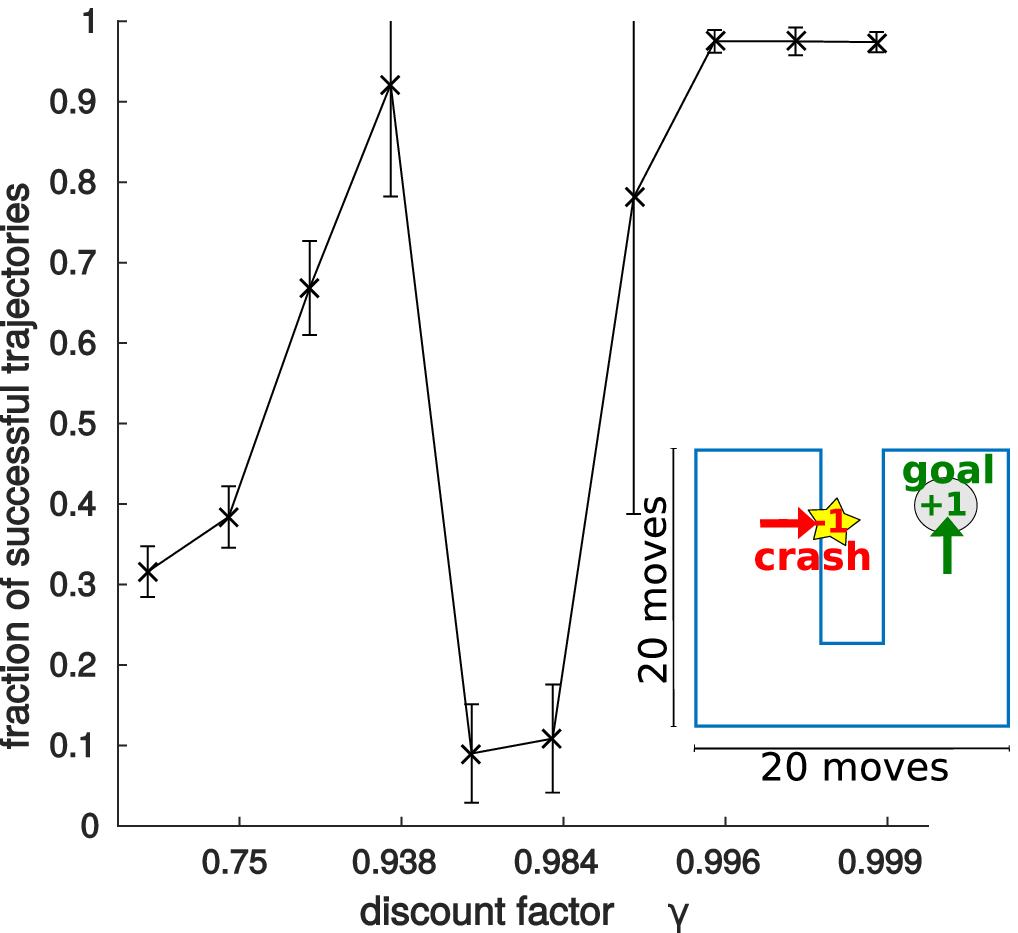}
	\hfill
	\includegraphics[width=0.485\textwidth]%
		{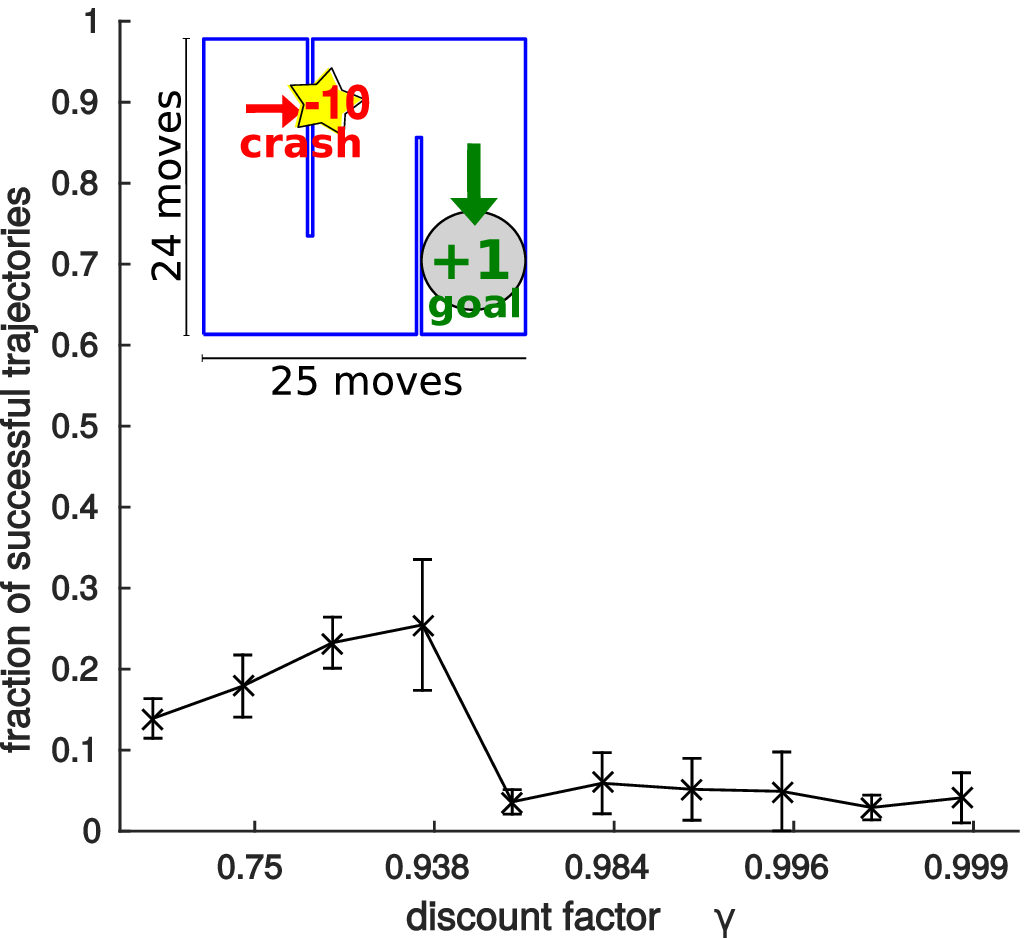} 
	\caption{ \label{fig:lspi_divergence}
		Navigation performance of policies, 
		learned by LSPI	in two environments
		(see sketched layouts),
		for varying {\em discount factors} $\gamma$.
		Error bars indicate mean and standard deviation 
		of the {\em fraction of successful test-trajectories} 
		(starting at random positions)
		over 10 random-walk training sets with 50000 samples each.
		The agent can either move forward or rotate 45$^\circ$
		left or right (i.e.~3 actions).
		Reaching the goal area is rewarded (+1) and
		crashing into a wall is punished (-1 or -10).
	}
\end{figure}

In difference to online algorithms,
which are often guaranteed to converge in the limit 
of an infinite training sequence \citep[e.g.][]{Sutton09},
batch learning has long been known to 
be vulnerable to the choice of training sets 
\citep{Tsitsiklis97,Bertsekas07}.
Depending on the batch of training samples at hand,
an RL algorithm can either converge 
to an almost optimal or to an arbitrarily bad policy.
In practice, this depends strongly (but not predictably)
on the {\em discount factor} $\gamma$.
For example, in Figure \ref{fig:lspi_divergence}
we demonstrate that policies learned by 
{\em least-squares policy iteration} \citep[LSPI,][]{Lagoudakis03}
yield very different performances 
when the discount factor $\gamma$ is varied
(experimental details can be found in Section \ref{sec:experiments}).
The left plot shows an unpredictable drop 
in performance for a simple navigation experiment 
with continuous states and discrete actions,
and the right plot the failure of LSPI to learn
a suitable policy for a more complicated environment.
The young discipline of deep RL has not yet reported 
effects like these, but it is reasonable to assume
that they happen in batch algorithms with 
more sophisticated architectures as well.

% ..............................................................................
% ..............................................................................

Most authors attribute this instability to 
a lack of convergence guarantees in off-policy batch value estimation
\citep[see][for an overview]{Dann14}.
But the distribution of training samples in the batch 
may also have a profound impact 
on the policy improvement in approximate RL.
For example, \citet{Perkins02} show for an algorithm similar to LSPI, 
that the {\em greedy policy improvement} can cause the instability
shown in Figure \ref{fig:lspi_divergence}.
Although their analysis does not carry over to LSPI\footnote{
	\citet{Perkins02} use open-ended on-policy online value estimation. 
	Training samples are drawn every time the policy is improved
	and errors on observed samples can thus average out over time.
}, they show that a sequence of non-deterministic policies
converge reliably when they are changed {\em slowly enough}.
{\em Conservative policy iteration} \citep[CPI,][]{Kakade02} 
follows a similar line of thought and slows down
the policy improvement considerably to guarantee convergence\footnote{
	In CPI, the next policy $\pi_{i+1}$ is a combination 
	of the previous policy $\pi_i$ 
	and the greedy policy $\pi_{i+1}^*$,
	i.e., $\pi_{i+1} = (1-\alpha) \pi_i + \alpha \,\pi_{i+1}^*$.
	CPI converges for small $\alpha \in [0,1]$.
	In SPI the update rate $\alpha$ is determined
	by maximizing a lower bound on the policy improvement,
	which converges much faster than CPI.
}. {\em Safe policy iteration} \citep[SPI,][]{Pirotta13}
extends this concept by determining the speed of change 
through a lower bound on the policy improvement.
The algorithm improves convergence speed significantly,
but is computationally expensive even in finite state spaces.
Other approaches suggest
an actor-critic architecture to 
avoid oscillations \citep{Wagner11}
or optimize a parameterizable softmax-policy directly 
\citep{Azar12}.

In this paper we evaluate the idea of \citet{Perkins02} 
empirically with LSPI in continuous navigation tasks.
Surprisingly, we find that the {\em stochasticity} 
of the improved policy stabilizes the solution, 
rather than the slowness of policy change.
This requires only a small modification to the policy improvement scheme.
Although our approach is a heuristic and 
theoretically not as well-grounded 
as the above algorithms, it is fast, simple to implement,
and can be applied to most algorithms used in deep RL.

% ==============================================================================
\section{Non-Deterministic Policy Improvement}
In this paper we consider tasks with 
continuous state space $\Set X$ and discrete\footnote{
	The extension to continuous action spaces is straight forward,
	but requires to compute an integral for each 
	application of the policy improvement operator 
	$\policyop{\beta}[f|q](x) = \int \pi_\beta^q(a | x) \, f(x,a) \, da$.
} action space $\Set A$.
A non-deterministic policy 
$\pi(a|x) \in [0,1]\,, \forall a \in \Set A\,, 
\sum_{a' \in \Set A} \pi(a'|x) = 1 \,, \forall x \in \Set X\,,$
can be evaluated by any algorithm to estimate 
the corresponding Q-value function $q : \Set X \times \Set A \to \R$.
To converge to the optimal policy,
the policy $\pi$ must also be {\em improved},
either during Q-value estimation or in an additional step.
The improvement in a state $x \in \Set X$ 
usually chooses the action $a \in \Set A$ that 
maximizes the current Q-value estimate $q(x,a)$.
Instead of this greedy improvement, 
we propose to produce an improved non-deterministic policy.
Examples are {\em softmax} $\pi_\beta^q$ or 
{\em $\epsilon$-greedy} $\pi_\epsilon^q$ policies\footnote{
	The softmax is also called 
	the {\em Boltzmann} or the {\em Gibbs policy}.
	Note the similarities to the policies of \citet{Wagner11}
	and \citet{Azar12}, which both implement a softmax 
	based on the optimized function.
}, that is, $\forall a \in \Set A, \forall x \in \Set X$:
$$
	\pi_\beta^q(a|x) = \frac{\exp\big(\beta \,  q(x,a)\big)}
		{\sum\limits_{a' \in \Set A} \exp\big(\beta \,  q(x,a')\big)}
	\quad\;\; \text{or} \quad\;\;
	\pi_\epsilon^q(a|x) = \epsilon \, \smallfrac{1}{|\Set A|} 
		+ \left\{ \begin{array}{cl}
			1-\epsilon &\kern-1.5ex, \text{if}\, 
				a = \argmax\limits_{a' \in \Set A} q(x,a') \\
			0 &\kern-1.5ex, \text{otherwise}
		\end{array} \right. \hspace{-2.5mm}\,.
$$
Existing algorithms can be adapted by identifying the 
greedy policy improvement operator $\policyop{*}$
and replacing it with the non-deterministic $\policyop{\beta}$,
that is, for functions $f,q: \Set X \times \Set A \to \R$:
\begin{equation} \nonumber
	\policyop{*}[f|q](x) \;=\; 
		f\big(x, \argmax_{a \in \Set A} q(x,a) \big)
	\quad\; \Rightarrow \quad\;
	\policyop{\beta}[f | q](x) \;=\; 
		\sum_{a \in \Set A} \pi_\beta^q(a | x) \, f(x,a) \,,
	\qquad \forall x \in \Set X \,.
\end{equation}
Here $\beta \in [0,\infty)$ denotes 
the inverse {\em stochasticity} of the operator.
For example, a non-deterministic version of
the TD-error $\delta_t$ in Q-learning 
for the observation $(x_t,a_t,r_t,x_{t+1})$ is 
$\delta_t = r_t + \gamma \, \policyop{\beta}[q|q](x_{t+1}) - q(x_t,a_t)$,
and the matrix $\mat A \in \R^{m \times m}$, 
which has to be inverted during non-deterministic
{\em least-squares temporal difference learning} 
\citep[LSTD,][used by LSPI]{Bradtke96}, 
would be computed from a training batch $\{ x_t,a_t,r_t \}_{t=0}^n$ by
$$ 
	A_{ij} \;\;=\;\; \smallfrac{1}{n}
	\smallsum{t=0}{n-1} \phi_i(x_t,a_t) \Big(
	\phi_j(x_t, a_t) - \gamma \policyop{\beta}[\phi_j | q](x_{t+1}) \Big) \,,
	\qquad \forall i,j \in \{1,\ldots,m\} \,.
$$
Softmax policies use more information than $\epsilon$-greedy
and are in most situations the better choice.
However, the stochasticity of the softmax 
depends strongly on the differences between Q-values.
Far away from the reward, Q-values can become very similar
and softmax policies become almost uniform distributions.
The level of stochasticity 
turns out to be the most reliable stabilizer for LSPI, 
and we used in our experiments (see Section \ref{sec:experiments})
{\em normalized Q-values} $\bar q$
for non-deterministic policy improvement $\policyop{\beta}[f|\bar q]$.
This normalizes the stochasticity for all states
by normalizing the difference between Q-values,
that is, $\forall x \in \Set X, \forall a \in \Set A$:
$$
	\bar q(x,a) = \frac{q(x,a) - \mu(x)}{\sigma(x)} \,, \quad
	\mu(x) = \smallfrac{1}{|\Set A|} \smallsum{a' \in \Set A}{} q(x,a') \,, \quad
	\sigma(x) = \sqrt{\smallfrac{1}{|\Set A|} \smallsum{a' \in \Set A}{}
		\big(q(x,a')\big)^2 - \big(\mu(x)\big)^2 } \,.
$$

% ==============================================================================
\section{Experiments} \label{sec:experiments}

We evaluated the effects of non-deterministic policy improvement 
at the example of a simple navigation experiment 
in an U- and a S-shaped environment 
(see inlays of Figure \ref{fig:lspi_divergence}).
The three dimensional state space $\Set X$ consisted of the agent's
two-dimensional position and its orientation.
The action space $\Set A$ contained 3 actions:
a forward movement and two 45$^\circ$ rotations.
Crashing into a wall stopped movement
and it would take the agent between 20 and 25 unimpeded moves to
traverse the environment in one spatial dimension.
Reaching the goal area (gray circle in the inlays)
yielded a reward of +1 and crashes 
incurred a punishment of -1 in the U-shaped 
and -10 in the S-shaped environment.
To represent the Q-value function, 
we chose a {\em Fourier basis} \citep{Konidaris11} and constructed
1500 basis functions over the space of states and actions.
The bases contained all combinations of:
10 cosine functions (including a constant) for each spatial dimension;
a constant, 2 cosine and 2 sine functions for the orientation;
and 3 discrete Kronecker-delta functions for the actions.
Irrespective their policy improvement,
policies were evaluated greedily 
to remain comparable.
Performance was measured in {\em fraction of successful trajectories},
which we estimated by running the greedy 
policy from 200 random starting positions/orientations.
Successful trajectories reach the goal 
within 100 actions without hitting a wall.

% ..............................................................................
\begin{figure}[p!]
	\includegraphics[width=0.485\textwidth]%
		{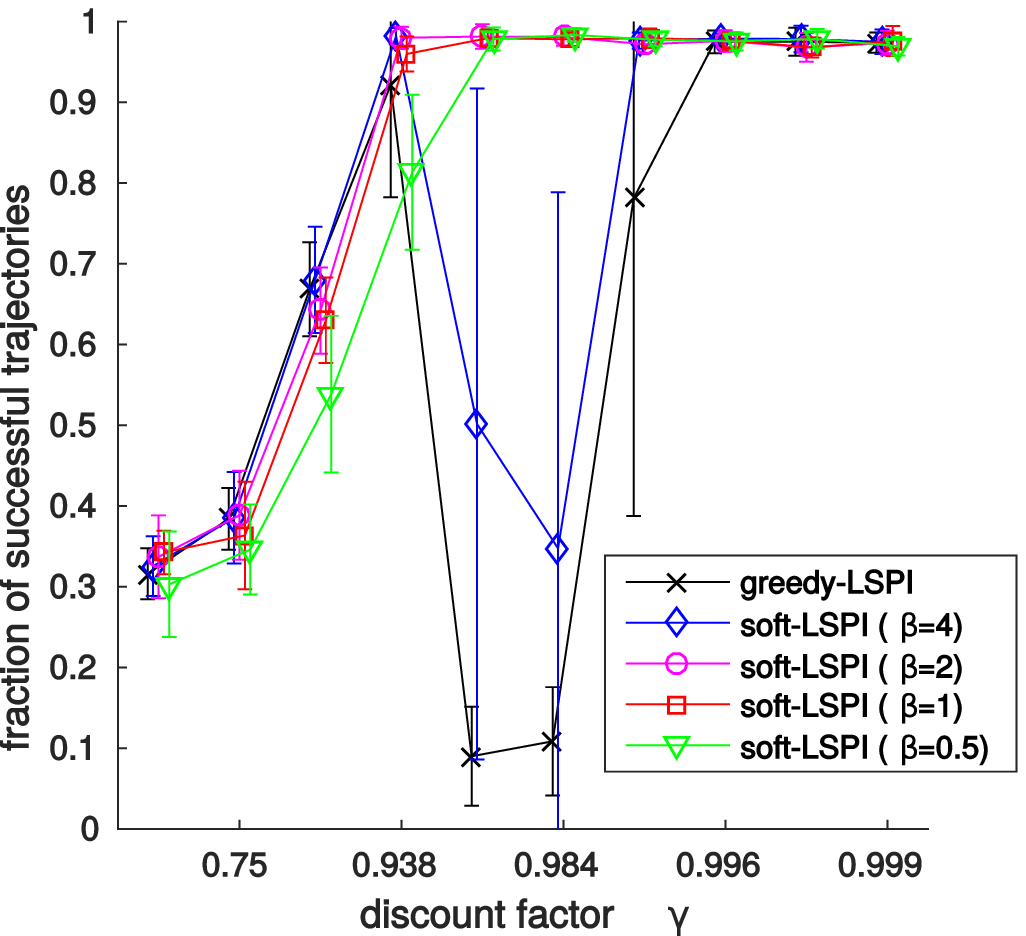}
	\hfill
	\includegraphics[width=0.485\textwidth]%
		{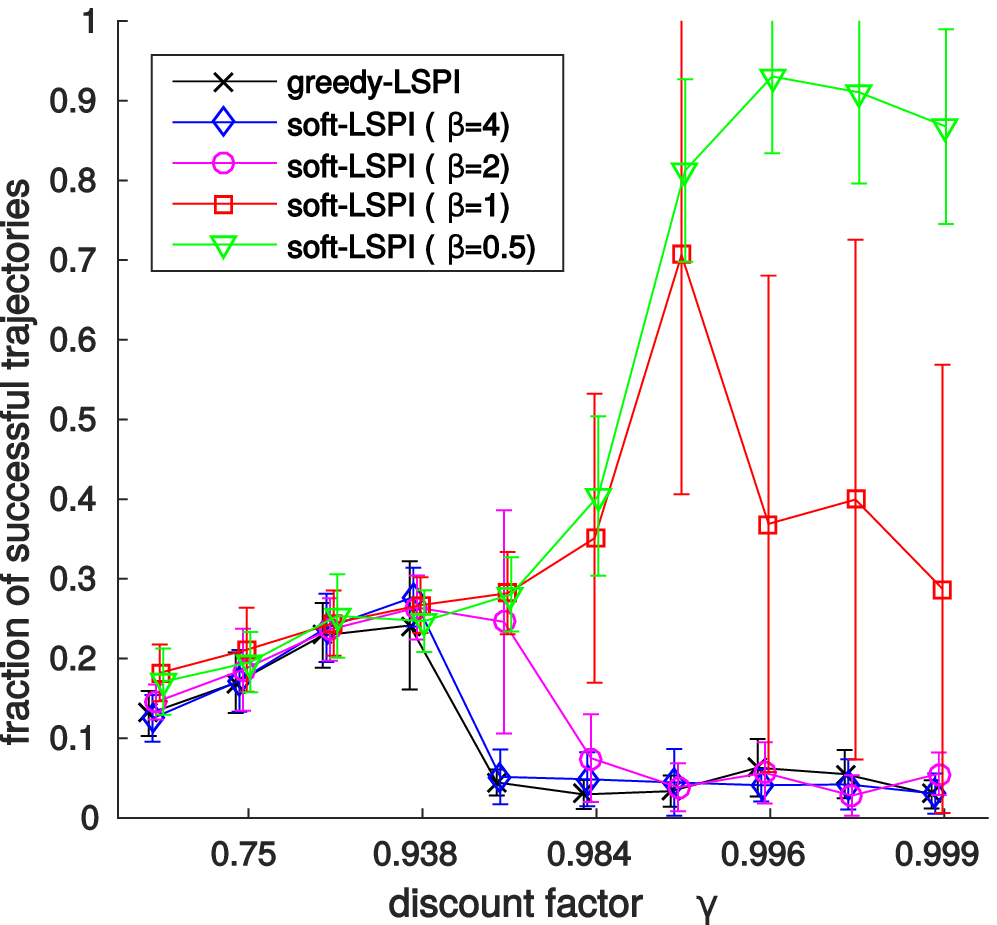} 
	\caption{ \label{fig:lspi_hard_soft}
		LSPI with greedy and softmax policy improvement,
		compared in the navigation tasks of Figure \ref{fig:lspi_divergence}.
		Large standard deviations are usually caused 
		by a mixture of excellent and horrible policies.
		We therefore call these regimes \qu{instable}.
		Stochastic improvements (with small $\beta$, e.g.~green triangles) 
		decrease performance for small $\gamma$,
		but stabilize convergence for large $\gamma$ significantly.
	}
\end{figure}
% ..............................................................................

% ..............................................................................
\begin{figure}[p!]
	\includegraphics[height=6.4cm]%
		{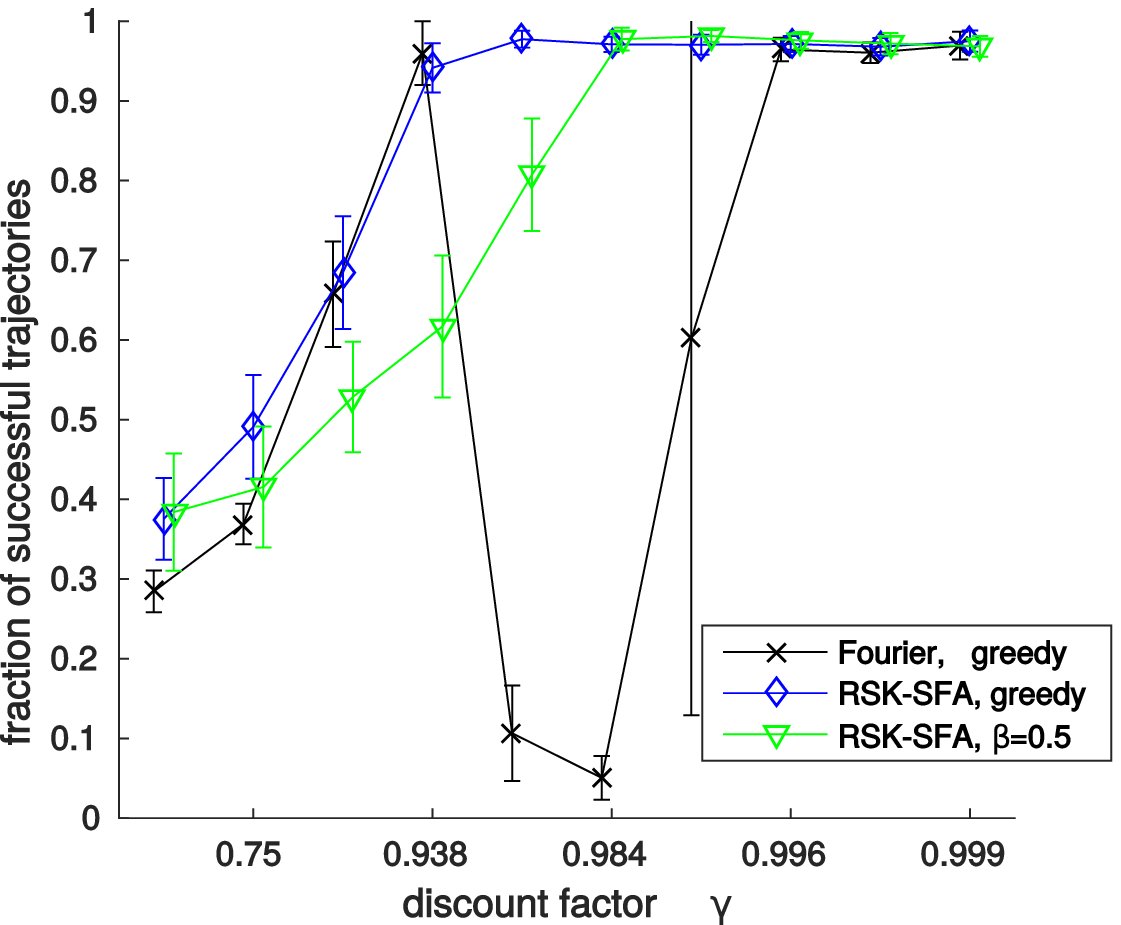}
	\hfill
	\includegraphics[height=6.4cm]%
		{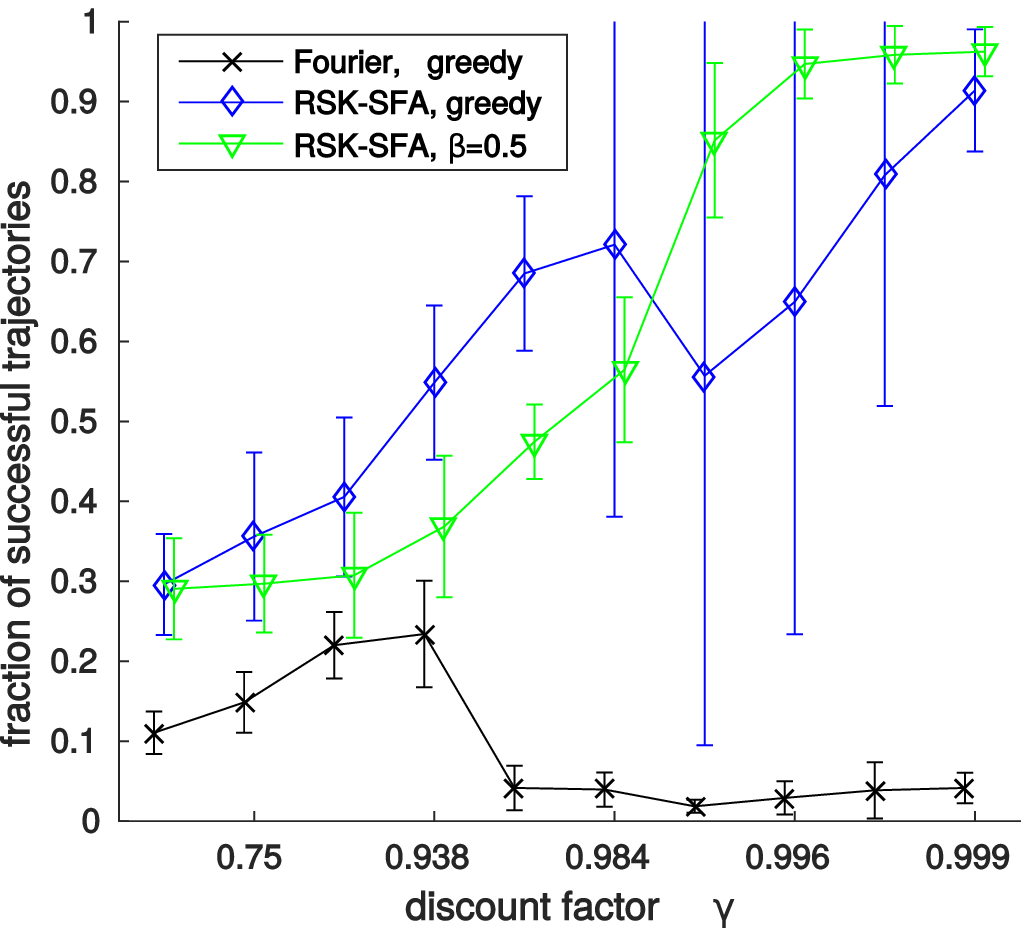} 
	\caption{ \label{fig:lspi_representation}
		LSPI policies based on different representations 
		in the navigation tasks of Figure \ref{fig:lspi_divergence}.
		Better representations (here RSK-SFA) generally improve performance,
		but non-deterministic policy improvement 
		is still needed to stabilize LSPI in complex tasks 
		(e.g., right plot).
		Also note the pronounced trade-off between $\beta$ and $\gamma$.
	}
\end{figure}
% ..............................................................................

% ------------------------------------------------------------------------------
\subsection{Non-Deterministic Policy Improvement}
We started out to test the idea of \citet{Perkins02}
for LSPI by using non-deterministic policy improvement 
(soft-LSPI) with slowly growing inverse stochasticity $\beta$
\citep[similar to {\em simulated annealing},][]{Haykin98}.
However, we observed that the annealing process 
itself did not improve the learned policy.
The performance was always comparable 
to soft-LSPI with the annealing's 
final stochasticity $\beta$ (not shown here).

Figure \ref{fig:lspi_hard_soft} plots the performance
of greedy-LSPI and soft-LSPI (with constant stochasticity $\beta$)
for varying discount factors $\gamma$.
In the face of sparse rewards, 
$\gamma$ determines how far that reward is propagated,
before it is drowned in inevitable approximation errors.
Low $\gamma$ yields policies 
that are only correct close to the reward,
and have therefore a bad performance.
On the other hand, $\gamma$ close to 1 
can lead to nearly optimal policies everywhere,
but performance is strongly affected 
by the instability investigated in this paper.
Note that the large standard deviations in both plots 
stem from {\em some} training sets producing near optimal, 
while others producing nonsensical policies.
For this reason we refer to these regimes as \qu{instable}.
First, one can observe that increasing stochasticity
(lower $\beta$) drastically stabilizes the soft-LSPI policies.
Secondly, note that there seems to be a trade-off
between inverse stochasticity $\beta$ and discount factor $\gamma$.
Low $\beta$ reduces performance while increasing stability, 
but in the left plot the performance with low $\beta$
becomes near optimal for larger $\gamma$, too.
It appears therefore that instabilities can generally be counteracted 
by simultaneously lowering $\beta$ and raising $\gamma$.

% ------------------------------------------------------------------------------
\subsection{Stabilization by Representation}
So far the above instabilities have only been demonstrated for LSPI.
One could argue that more sophisticated approaches 
must not be affected in the same way. 
In deep neural networks, for example, 
the lower layers may provide 
a {\em representation} of the state-action space
that stabilizes policy improvement.
We want to investigate this by choosing basis functions, 
which are known to represent value functions well.
\citet{Boehmer13a} show that features learned by non-linear
{\em slow feature analysis} \citep[SFA,][]{Wiskott02}
approximate an optimal encoding for value functions 
of all tasks in the same environment\footnote{
	Strictly speaking, 
	this holds only for values of the 
	{\em sampling policy} of the training data.
	However, SFA features are reported to work well 
	with LSPI for random-walk training sets \citep{Boehmer13a}.
}. We used {\em regularized sparse kernel SFA} \citep[RSK-SFA,][]{Boehmer12} 
with Gaussian kernels to learn such features from the training data.
Figure \ref{fig:lspi_representation} shows the results in 
comparison with the trigonometric 
Fourier basis functions introduced above.
Using the SFA representation completely avoided instability 
for the simpler task in the left plot (blue diamonds).
The performance improves in the S-shaped environment too, 
but the large standard deviations indicate that here greedy LSPI
is not very stable for large discount factors $\gamma$.
Soft-LSPI with a low $\beta$ (green triangles) 
stabilizes the solution, though.
Using a deep architecture may therefore {\em reduce} instability, 
but will probably {\em not remove} it all together.
Nonetheless, our results suggest that non-deterministic 
policy improvement should be able to stabilize deep architectures, too.

% ==============================================================================
\section{Conclusion}
We have shown that (at least) LSPI can become {\em instable}
in some unpredictable regimes of the discount factor $\gamma$.
Here small differences in the  training set 
can lead to large differences in policy performance.
It is not exactly clear why solutions become instable,
but we show that learned policies can be stabilized 
by using a non-deterministic policy improvement scheme.
All presented experiments became significantly 
more stable by increasing stochasticity $\frac{1}{\beta}$ 
and discount factor $\gamma$ at the same time.
Future works may extend our approach by adjusting 
both parameters during policy iteration 
\citep[like in SPI,][]{Pirotta13}.
Better representations of the state-action space
have also improved stability to some extend.
More sophisticated approaches (like deep RL)
learn these representations implicitly in 
their lower layers and may therefore be more stable than LSPI.
Nonetheless, instabilities {\em will} probably occur,
and non-deterministic policy improvement can 
most likely be employed to stabilize the learned policy
in deep RL, too.

In conclusion, when success or failure of learned policies 
depends crucially on the training set (e.g.~during cross-validation),
one should consider a non-deterministic policy improvement scheme.
The scheme presented in this paper is 
computationally cheap, easy to implement,
and can be fine-tuned with the inverse stochasticity $\beta$.

\newpage
\acks{We thank the anonymous reviewers, 
who pointed us in the direction of CPI. 
This work was funded by the {\em German science foundation} (DFG)
within SPP 1527 {\em autonomous learning}.}

% ==============================================================================
% Acknowledgements should go at the end, before appendices and references

%\acks{We would like to acknowledge funding agencies. }

% Manual newpage inserted to improve layout of sample file - not
% needed in general before appendices/bibliography.

%\newpage
% ==============================================================================
\vskip 0.2in
{%\tiny 
	\bibliography{bibliography} 
}

\end{document}